\documentclass[lettersize,journal]{IEEEtran}
\usepackage{amsmath,amsfonts}
\usepackage{algorithmic}
\usepackage{algorithm}
\usepackage{array}
\usepackage[caption=false,font=normalsize,labelfont=sf,textfont=sf]{subfig}
\usepackage{textcomp}
\usepackage{stfloats}
\usepackage{url}
\usepackage{verbatim}
\usepackage{graphicx}
\usepackage{cite}
\usepackage{cleveref}
\usepackage{booktabs} 
\usepackage{multirow} 
\usepackage[table,xcdraw]{xcolor}
\usepackage{makecell} 

\usepackage{pifont}
\usepackage{soul}
\newcommand{\cmark}{\ding{51}}%
\newcommand{\xmark}{\ding{55}}

\begin{document}

\title{Human-Centric Video Anomaly Detection Through Spatio-Temporal Pose Tokenization and Transformer}

\author{Ghazal Alinezhad Noghre,~\IEEEmembership{Student Member,~IEEE,}
        Armin Danesh Pazho,~\IEEEmembership{Student Member,~IEEE,}
        Hamed Tabkhi,~\IEEEmembership{Member,~IEEE}}



\maketitle

\begin{abstract}
Video Anomaly Detection (VAD) presents a significant challenge in computer vision, particularly due to the unpredictable and infrequent nature of anomalous events, coupled with the diverse and dynamic environments in which they occur. Human-centric VAD, a specialized area within this domain, faces additional complexities, including variations in human behavior, potential biases in data, and substantial privacy concerns related to human subjects. These issues complicate the development of models that are both robust and generalizable. To address these challenges, recent advancements have focused on pose-based VAD, which leverages human pose as a high-level feature to mitigate privacy concerns, reduce appearance biases, and minimize background interference. In this paper, we introduce SPARTA, a novel transformer-based architecture designed specifically for human-centric pose-based VAD. SPARTA introduces an innovative Spatio-Temporal Pose and Relative Pose (ST-PRP) tokenization method that produces an enriched representation of human motion over time. This approach ensures that the transformer's attention mechanism captures both spatial and temporal patterns simultaneously, rather than focusing on only one aspect. The addition of the relative pose further emphasizes subtle deviations from normal human movements. The architecture's core, a novel Unified Encoder Twin Decoders (UETD) transformer, significantly improves the detection of anomalous behaviors in video data. Extensive evaluations across multiple benchmark datasets demonstrate that SPARTA consistently outperforms existing methods, establishing a new state-of-the-art in pose-based VAD.
\end{abstract}


\section{Introduction}
\label{sec:intro}
\IEEEPARstart{V}{ideo} Anomaly Detection (VAD) is a rapidly growing area within Computer Vision that focuses on automatically identifying unusual events or behaviors in video sequences \cite{ren2021deep,patrikar2022anomaly, abbas2022comprehensive, ramachandra2020survey}. This technology has a wide array of practical applications, including smart surveillance \cite{pazho2023ancilia, patrikar2022anomaly}, traffic monitoring \cite{zhao2023unsupervised, yu2022deep, Pazho_2024_CVPR}, and healthcare \cite{nanda2022soft}. An established subset of VAD is human-centric anomaly detection, which specifically targets recognizing atypical human behaviors.

\begin{figure*}[]
    \centering
    \includegraphics[clip,trim={16 23 20 17},width=0.8\linewidth]{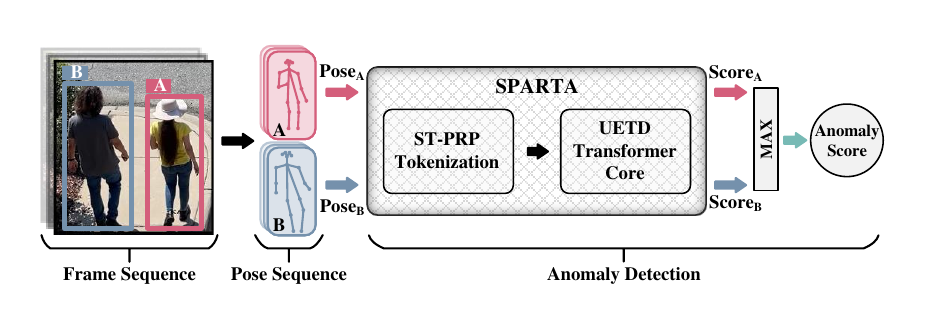}
    \caption{A conceptual overview of SPARTA. SPARTA assigns higher scores to anomalous pose sequences. The final frame score is the maximum score of all individuals in the scene.}
    \label{fig:intro}
\end{figure*}

The complexity of this task stems from its open-set nature, where an immense variety of normal and abnormal human behaviors can occur in real-world situations. For example, anomalies could include someone falling, engaging in a physical altercation, or causing unusual congestion in public spaces \cite{zhu2022towards}. The core challenge is the unpredictability and diversity of these events. Traditional supervised training methods, which depend on datasets that may not cover the full spectrum of possible anomalies, struggle with generalizability \cite{alinezhad2023understanding}. This is because anomalies, by definition, are often unknown and unexpected.

To overcome these challenges, the field is increasingly embracing self-supervised learning approaches. These innovative methods enhance the performance of anomaly detection models by learning from normal data without needing explicit labels or fine-grained categories for anomalies \cite{markovitz2020graph, hirschorn2023normalizing, morais2019learning}. In essence, self-supervised models develop an understanding of normal human behavior patterns. Consequently, any deviation from these learned patterns is identified as an anomalous event. In this paper, we adopt the self-supervised learning approach, aligning with recent research trends, to address the inherent challenges in human-centric anomaly detection.

Regardless of the training methodology employed, human-based VAD is primarily categorized into two strategies: pixel-based \cite{wang2023memory, zaheer2022generative, georgescu2021anomaly, ristea2022self, wang2022video} and pose-based \cite{morais2019learning, markovitz2020graph, jain2021posecvae, rodrigues2020multi, zeng2021hierarchical, yu2023regularity} methods. For video processing and training, a general technique involves analyzing the pixel data of frames over time. This holds true for VAD as well, where pixel-based approaches examine the raw pixel values in video frames to identify anomalies, leveraging the fine granularity provided by the pixels in each frame. Nonetheless, focusing specifically on human behaviors, the examination of every pixel can result in the analysis of excessive, redundant pixels, introducing undesirable noise into the system \cite{hirschorn2023normalizing}. Such noise can range from relatively harmless disturbances like background changes to more critical issues like demographic attributes (skin color, clothing, gender, etc.), potentially inducing biases within the system. While such noise and biases might not show their effect through available metrics and datasets, they becomes critical in the deployment of the models in the real world.

Pose-based methods have been developed to mitigate these effects. These methods concentrate on the poses of individuals within the scene, offering a more refined understanding of human movements \cite{yu2023regularity}. By prioritizing human poses, these methods not only enhance privacy but also reduce demographic biases and exhibit enhanced resilience against background disturbances, thereby proving their efficacy in diverse real-world applications \cite{hirschorn2023normalizing, yu2023regularity, alinezhad2023understanding}. Consequently, this study delves deeper into human pose analysis and leverages it for VAD.

Given the need for self-supervised training and the sequential nature of input pose data transformers emerge as an attractive choice of architecture. Transformers have revolutionized fields such as natural language processing and time series analysis \cite{li2022self, vaswani2017attention}, are ideally suited for self-supervised learning frameworks \cite{liu2023survey}, effectively exploiting the sequential patterns in input data. They excel at capturing long-range dependencies \cite{khan2022transformers, sanford2024representational}, a crucial attribute for identifying intricate patterns in extensive sequences, such as those encountered in VAD. The challenge, however, lies in adapting the Transformer's robust attention mechanism to the nuanced requirements of Computer Vision. Here, we aim to bridge this gap by developing a novel tokenization strategy that not only leverages this mechanism but also aligns with the human pose data utilized in our VAD model. This synergy aims to enhance our model's effectiveness in identifying anomalies in video data while preserving the privacy and bias reduction benefits inherent to pose-centric techniques.


This paper introduces SPARTA, a novel non-autoregressive transformer-based model with an innovative spatio-temporal tokenization approach for pose-based human-centric anomaly detection. The non-autoregressive design of SPARTA enables the simultaneous generation of output tokens, crucial for the time-sensitive nature of anomaly detection. \cref{fig:intro} provides an abstract overview of the SPARTA system. SPARTA features two innovative components: the Spatio-Temporal Pose and Relative Pose (ST-PRP) tokenization and the Unified Encoder Twin Decoders (UETD) transformer core. The ST-PRP tokenization is designed to maximize the self-attention capabilities of the transformer, creating a new paradigm in pose tokenization for a range of advanced pose-based tasks. The UETD core, with its Future Target Decoder (FTD) and Current Target Decoder (CTD), processes tokens and calculates anomaly scores. This architecture combines a single unified encoder with two decoder heads, each tailored for specific operational goals.

SPARTA, with just 0.5 million parameters, introduces a self-supervised approach to anomaly detection that sets a new standard for performance across multiple benchmark datasets, including ShanghaiTech \cite{liu2018future}, HR-ShanghaiTech \cite{morais2019learning}, Charlotte Anomaly Dataset \cite{danesh2023chad}, and Northwestern Polytechnical University Campus \cite{cao2023new}, achieving State-of-the-Art (SOTA) average Area Under the Receiver Operating Characteristic Curve (AUC-ROC) score of $75.87\%$. Additionally, crucial for real-world applications, SPARTA maintains an average Equal Error Rate (EER) of $0.29$, demonstrating SOTA balance between false negatives and false positives on these datasets.

This paper presents the following contributions:

\begin{itemize}
    \item Introducing the Spatio-Temporal Pose and Relative Pose (ST-PRP) tokenization as a novel approach for tokenization of human pose for anomaly detection and showcasing its benefits through extensive ablation study.
    \item Introducing SPARTA, the combination of a novel non-autoregressive Unified Encoder Twin Decoders (UETD) transformer and the ST-PRP tokenization, featuring Current Target Decoder (CTD) and Future Target Decoder (FTD) for self-supervised human anomaly detection.
    \item Demonstrating the superior accuracy and generalizability of SPARTA through comparison with not only SOTA pose-based approaches but also pixel-based approaches across four benchmark datasets.
\end{itemize}

\section{Related Works}
\label{sec:related}

The field of anomaly detection has evolved, adapting to scientific advancements, particularly within the realm of Artificial Intelligence (AI) \cite{patrikar2022anomaly, zhu2020video}. The trend started from handcrafted methods \cite{cocsar2016toward, cheng2015gaussian, yuan2014online} with approaches utilizing algorithms such as histogram of optical flow \cite{kaltsa2015swarm}. With the advancements in deep neural networks, anomaly detection took a leap forward utilizing the learning capabilities of Convolutional Neural Networks (CNNs) \cite{kong2021real, sarker2021semi, cheng2020securead, sun2020adversarial, li2020spatiotemporal}. To learn more features, approaches also started adopting Deep Neural Networks (DNNs) \cite{sabokrou2017deep, suresha2020study, georgiou2020survey, kim2021deep}. Long Short-Term Memory was another advancement known for handling time series data. They were widely used in video analysis and surveillance anomaly detection \cite{ergen2019unsupervised, ullah2021cnn, ullah2021efficient, sabih2022crowd, asad2021multi}. Generative Adversarial Networks (GANs) represent another advanced methodology that numerous researchers have adopted for anomaly detection \cite{jackson2021svd, saypadith2021approach, yang2021bidirectional, zhang2021generative, dong2020dual}. The most recent approaches have started to leverage transformer architectures \cite{ullah2023transcnn, chen2023mgfn, li2022self, zhang2022weakly, huang2022weakly, sun2022transformer, madan2023self, wu2022self} owing to their versatility and deep understanding capabilities enabled by the self-attention module \cite{vaswani2017attention}.

There are two main categories for self-supervised human-centric anomaly detection. First are Pixel-based approaches \cite{ barbalau2023ssmtl++, madan2023self, ristea2022self, li2022scale, yang2022dynamic} with various subgroups such as Spatio-Temporal Jigsaw Puzzle \cite{wang2022video}, and Multi-Task Design \cite{georgescu2021anomaly}. Inherently these methods can have an internal bias towards the appearance features of the individuals in the scene as well as high sensitivity towards background noise \cite{buet2022towards, steed2021image}. Like pixel-based approaches, pose-based algorithms can also be divided into multiple subgroups such as methods that use Spatio-Temporal Graph Convolution \cite{luo2021normal}, Multi-Scale Prediction \cite{rodrigues2020multi}, and Hierarchical Prediction \cite{zeng2021hierarchical}.

Normal Graph \cite{luo2021normal} leverages Spatio-Temporal Graph Convolution. When the model, trained only on normal behaviors, predicts future movements that greatly differ from actual movements, these disparities indicate anomalous behavior. \cite{rodrigues2020multi} employs future and past prediction modules to enhance the accuracy of their anomaly detection model through multi-scale past/future prediction. \cite{zeng2021hierarchical} propose a hierarchical prediction-based method, utilizing three branches to predict pose, trajectory, and motion vectors. MPED-RNN \cite{morais2019learning} and \cite{li2022human} employ encoder-decoder structures for anomaly detection utilizing recurrent neural networks. MemWGAN-GP \cite{li2023human} leverages generative adversarial networks by employing a dual-head decoder structure and upgrading it with a modified version of the Wasserstein Generative Adversarial Network (WGAN-GP) \cite{pmlr-v70-arjovsky17a}. GEPC \cite{markovitz2020graph} combines a spatio-temporal graph autoencoder with a clustering layer to assign soft probabilities to input pose segments, serving as an anomaly score. \cite{yu2023regularity} incorporates transformers into anomaly detection. This method combines an encoder-only transformer with a simple linear layer as a reconstruction head. STG-NF \cite{hirschorn2023normalizing} proposes a model that uses normalizing flows to map human pose data distribution to a fixed Gaussian distribution, leveraging spatio-temporal graph convolution blocks. In contrast to all previous methodologies, our approach integrates novel spatio-temporal tokenization, as well as a newly introduced Unified Encoder Twin Decoders transformer core processing, pose through a combination of Current Target Decoder (CTD) and Future Target Decoder (FTD) to achieve the task of human-centric video anomaly detection.

\begin{figure*}[ht!]
    \centering
    \includegraphics[clip,trim={20 20 20 22},width=0.9\textwidth]{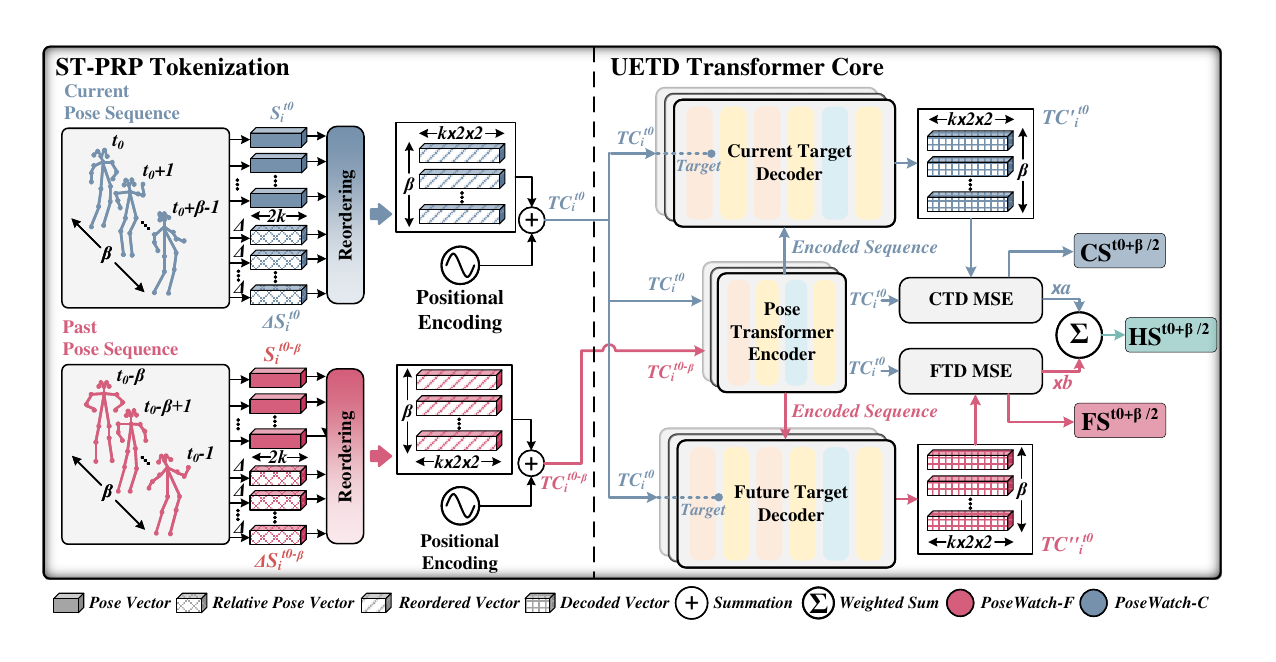}
    \caption{SPARTA architecture. ST-PRP tokenization reorders and prepares input pose sequences for being fed to the UETD transformer core. The UETD transformer core consists of a unified pose transformer encoder and twin decoders for CTD and FTD. The MSE loss of both CTD and FTD branches is used to calculate the Current Score (CS) and Future Score (FS), respectively. The average of the two scores is calculated to find the Hybrid Score (HS). Please note that $a$ and $b$ are constant multipliers both set to $0.5$ for calculating the HS. Red and blue represent SPARTA-F and SPARTA-C data flows respectively.}
    \label{fig:SPARTA}
\end{figure*}

\section{SPARTA}
\label{sec:SPARTA}



The architecture of SPARTA, depicted in \Cref{fig:SPARTA}, embodies two main components: the Spatio-Temporal Pose and Relative Pose (ST-PRP) tokenization, alongside the Unified Encoder Twin Decoders (UETD) transformer core. SPARTA leverages a shared encoder, a Current Target Decode (CTD), and a Future Target Decoder (FTD). The CTD and FTD branches provide complementary insights by capturing distinct patterns within the input sequences. This synergy enhances the overall model's robustness, as it minimizes the influence of individual branch errors through the aggregation of results from both branches. In the following subsections, we delve into the details of SPARTA.


\subsection{ST-PRP Tokenization}
\label{sec:st-prp}
The tokenization process aims to provide a rich and informative input sequence to the transformer model. We define the absolute pose sequence as follows:

\begin{equation}
\label{eq:abs_input}
    S_{i}^{t_0} = [P_{i}^{t_0}, P_{i}^{t_0+1}, P_{i}^{t_0+2}, \cdots, P_{i}^{t_0+\beta-1}] 
\end{equation}

where $S_{i}$ is the absolute pose sequence of person $i$, $P$ is pose data containing $(x, y)$ coordinates of the joints, $t_0$ is the starting frame of the sequence, and $\beta$ is the input window size. This will provide the model with basic information about the position of a person's joints in each sequence frame. However, a person's movement patterns through the sequence also reveal critical information for anomaly detection. To accentuate the global movement of humans through space, in addition to the absolute pose sequence, we also leverage the relative pose sequence shown in \Cref{eq:rel_input}. $\Delta S_{i}$ is constructed to highlight the overall movements of the subjects relative to the coordinates of the first pose of the current sequence as shown in \Cref{eq:rel_input_detail}.

\begin{equation}
    \label{eq:rel_input}
    \Delta S_{i}^{t_0} = [\Delta P_{i}^{t_0}, \Delta P_{i}^{t_0+1}, \Delta P_{i}^{t_0+2}, \cdots, \Delta P_{i}^{t_0+\beta-1}] 
\end{equation}
\begin{equation}
    \label{eq:rel_input_detail}
    \Delta P_{i}^{t} =  P_{i}^{t} -  P_{i}^{t_0}
\end{equation}

$\Delta P$ is relative pose data containing relative coordinates of the joints, $t_0$ is the starting frame of the sequence, and $\beta$ is the input window size.

\begin{figure}[]
    \centering
    \includegraphics[clip,trim={20 23 22 20},width=0.8\columnwidth]{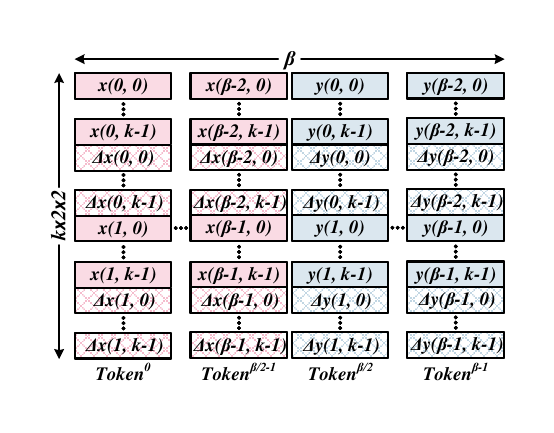}
    \caption{SPARTA Spatio-Temporal Pose and Relative Pose (ST-PRP) Tokenization Schema. $k$ is the number of keypoints, $\beta$ is the input window size, $\Delta$ shows relative coordinates and $x(t, k)$ and $y(t, k)$ are the coordinates of $k^{th}$ keypoint in time step $t$.}
    \label{fig:prp}
\end{figure}

The transformer's design exclusively employs inter-token self-attention, ignoring any intra-token attention mechanisms \cite{vaswani2017attention}. Consequently, the way attention is applied depends significantly on how the input data is tokenized. In order to utilize the full potential of the transformer self-attention module we introduce a Spatio-Temporal Pose and Relative Pose Tokenization (ST-PRP). After conducting extensive experiments with various tokenization strategies, as detailed in \Cref{sec:ablation}, we identified ST-PRP as the best-performing approach. ST-PRP tokenization, as depicted in \Cref{fig:prp}, employs $\beta$ tokens, each with dimensions of $k \times 2 \times 2$. The initial $\beta/2$ tokens are dedicated to $x$ coordinates, and the latter half pertains to $y$ coordinates. Each token encapsulates both the absolute and relative values of either $x$ or $y$ coordinates of a pose in two adjacent frames. Spatial attention arises from the attention between keypoints and their $x$ and $y$ dimensions, whereas temporal attention is obtained by the frame number progression across tokens. Transformers, by design, do not inherently understand the order of input unless it's explicitly provided. The ST-PRP tokenization captures spatial and temporal relationships within and across frames, but it doesn't inherently indicate the sequence of tokens. Thus, we use the positional encoding strategy \cite{vaswani2017attention} to embed order into the input sequence and construct the input of the SPARTA Unified Encoder Twin Decoders (UETD) transformer core or $TC_i$.
Further insights into the significance and characteristics of relative pose utilization, along with a detailed analysis of various tokenization methods, through empirical evaluations, are presented in \Cref{sec:ablation}.

\subsection{UETD Transformer Core}

At the heart of SPARTA is the UETD module, which processes tokens generated by the ST-PRP through a dual-branch structure. This core component is trained in a self-supervised manner, enabling effective anomaly detection. The following sections will go into the details of each of these branches and their architecture.

\subsubsection{CTD Branch (SPARTA-C):} The CTD branch works on the basis that a network trained on the normal data samples in the training set will learn how to encode normal pose sequences to the latent space and generate current sequence with a relatively low Mean Squared Error (MSE) loss. However, if this model is given abnormal pose sequences since it has not seen such datapoints during the training, its generative ability is compromised leading to a relatively larger MSE loss indicating abnormal behavior.

The ST-PRP Tokenization output serves as the input for SPARTA-C, being treated as a sequential data stream. The SPARTA CTD branch has an encoder-decoder structure. We chose our design to use a non-autoregressive strategy to take advantage of the parallelization of the transformer's structure. In the CTD branch, we chose the target sequence of the decoder to be equal to the current sequence $TC_i^{t_0}$. Both the encoder and the decoder are chosen to have 12 heads. Considering the real-time nature of the anomaly detection task, we choose a minimal 4 number of layers with the feed-forward layer dimensions set to 64. Unlike most NLP tasks, we do not need to define a start token and end token for the sequences since the input and output sequences have fixed lengths. SPARTA does not use masking strategies for both the input and target sequences since these sequences are always available even in the inference time. The output of the CTD branch for the input $TC_i^{t_0}$ is: 

\begin{equation}
\label{eq:recon_output}
     TC_{i}^{'t_0} = [Token^{'0}_i, Token^{'1}_i, \cdots, Token^{'\beta -1}_i]
\end{equation}

where $Token^{'n}_i$ is the $n^{th}$ generated token of the $i^{th}$ person in the $t_0$ sequence $TC_{i}^{'t_0}$.

Finally, the MSE loss between the generated sequence $ TC{'t_0}_{i}$ and the input sequence $TC_i^{t_0}$ is used both as the training loss and calculating the CTD Score ($CS^{t_0+\beta/2}_{i}$) in the inference time.



\subsubsection{FTD Branch (SPARTA-F):} As illustrated in \Cref{fig:SPARTA}, the SPARTA FTD and CTD branches utilize a shared encoder. This encoder leverages an advanced understanding of pose dynamics and progression when trained for the CTD branch. Correspondingly, the FTD branch employs a decoder mirroring the architecture of the CTD, leveraging 12 heads and 4 layers. Although these decoders have identical architecture (denoted as 'Twin Decoders'), each of them is distinctively tasked. While both generate the current sequence, their inputs differ in timing with FTD's input lagging by one sequence step compared to CTD. Following the non-autoregressive strategy, for the input sequence of $TC_i^{t_0 - \beta }$, the target sequence of the decoder is the future sequence ($TC_i^{t_0}$ is considered the future sequence compared to the input of  $TC_i^{t_0 - \beta }$) in both the training and inferencing process. For the same reason as the CTD branch, We do not use any masking in the FTD branch either. The output of the FTD branch for the input $TC_i^{t_0-\beta}$ is:

\begin{equation}
\label{eq:pred_output}
     TC_{i}^{''t_0} = [Token_{i}^{''0 }, Token_{i}^{''1}, \cdots, Token_{i}^{''\beta-1}]
\end{equation}
where $Token^{''n}_i$ is the $n^{th}$ generated token of the $i^{th}$ person in the $t_0$ sequence $TC_{i}^{''t_0}$. During the training of this branch, the pose encoder parameters are frozen, and only the FTD decoder parameters are trained. The MSE loss between the generated sequence and the actual sequence is used both for the training process and calculating the FTD Score ($FS^{t_0+\beta/2}_{i}$) at inference time.




\subsubsection{SPARTA Hybrid (SPARTA-H):} In order to be able to capture all anomalous patterns detected by both the CTD and the FTD branches, we combine the scores from these branches to calculate the Hybrid Score or $HS_{i}^t$ of the $i^{th}$ person. We use a weighted sum strategy described in \Cref{eq:final_score}. Before combining the scores, we normalize them to ensure they are in the same range.
\begin{equation}
\label{eq:final_score}
    \begin{split}    
        HS_{i}^{t_{0}+ \frac{\beta}{2}} = &  0.5 \cdot Norm( CS_{i}^{t_{0}+ \frac{\beta}{2}}) \\
        & +  0.5 \cdot Norm ( FS_{i}^{t_{0}+ \frac{\beta}{2}} )
    \end{split}
\end{equation}

In the last step, we find the maximum anomaly score across all people available in the scene to find one score for each frame:

\begin{equation}
    \label{eq:frame_score}
    HS^{t_{0}+ \frac{\beta}{2}} = max_{i\in N} (HS_{i}^{t_{0}+ \frac{\beta}{2}})
\end{equation}
where $N$ is the set of available people in the frame. 

In addressing the real-time demands of anomaly detection, our commitment to minimal model complexity is evident, exemplified by only choosing 4 layers for both the encoder and twin decoders. Additionally, diverging from vision transformers employed in various computer vision tasks, our strategy involves tokenized poses with reduced dimensions, resulting in SPARTA-H only having 0.5 million parameters and an average end-to-end latency of 5.96 ms.

\section{Experimental Setup}
\label{sec:setup}

\subsection{Datasets}

Early datasets such as CUHK Avenue \cite{lu2013abnormal}, Subway \cite{adam2008robust}, and UCSD \cite{wang2010anomaly} have been foundational for VAD research. However, their limited scale has prompted the research community to adopt more complex and comprehensive recent datasets, which aligns with our focus on utilizing these modern datasets for experimentation.

\textbf{ShanghaiTech Campus (SHT) \cite{liu2018future}} dataset is the primary benchmark for human-centric video anomaly detection, offering over 317,000 frames from 13 scenes. It includes 274,515 normal training frames and 42,883 test frames with both normal and anomalous events in its unsupervised split. The dataset features unique anomalies as well as various lighting conditions and camera angles, with 130 abnormal events. In line with previous SOTA approaches \cite{markovitz2020graph, hirschorn2023normalizing, yu2023regularity}, AlphaPose \cite{li2019crowdpose} is utilized for pose extraction and tracking to ensure fair comparison.

\textbf{HR-ShanghaiTech (HR-SHT) \cite{morais2019learning}} represents a human-related adaptation of the SHT dataset. Notably, the only distinction lies in its exclusive focus on human-centric anomalies.


\textbf{Charlotte Anomaly Dataset (CHAD) \cite{danesh2023chad}} is a new large-scale high-resolution multi-camera VAD dataset with about 1.15 million frames, including 1.09 million normal and 59,172 anomalous frames. Unique for its detailed annotations, including bounding boxes and poses for each subject, CHAD offers a more challenging environment compared to SHT. The experiments are conducted on the unsupervised split. CHAD is selected since it sets a unified benchmark for pose-based anomaly detection by providing extracted poses to eliminate the variations in the final pose-based anomaly detection accuracy.

\textbf{Northwestern Polytechnical University Campus (NWPUC) \cite{cao2023new}} dataset encompasses 43 scenes, 28 classes of anomalous events, and 16 hours of video footage, making it a large datasets in its field. A notable feature of the NWPUC dataset is its inclusion of scene-dependent anomalies, where an event may be considered normal in one scene but abnormal in another. This dataset includes anomaly classes that are not specific to humans, presenting a disadvantage for pose-based anomaly detection methods. However, as a new and comprehensive benchmark, it was chosen for its broad applicability. This limitation affects all pose-based methods, ensuring fair comparisons.


\subsection{Metrics}
\label{sec:metrics}
\textbf{AUC-ROC} or the Area Under the Receiver Operating Characteristic Curve is used to evaluate the discriminative power of models for binary classification. It plots the True Positive Rate (TPR) against the False Positive Rate (FPR) at various thresholds. A higher AUC-ROC value indicates better model performance in class separation.



\textbf{EER} or the Equal Error Rate represents the point at which the False Positive Rate (FPR) and False Negative Rate (FNR) are equal. Owing to the valuable insights that EER provides for anomaly detection, several works such as \cite{li2023human, li2022human} report it, but it is not used as widely as AUC-ROC. EER finds a balancing point between both error rates, indicating an optimal trade-off between security (inferred from FNR) and usability (inferred from FPR). Notably, EER is not influenced by imbalanced data, which is crucial for anomaly detection problem. On its own, this metric is not informative enough to evaluate a model \cite{sultani2018real}, but in conjunction with AUC-ROC, it provides additional valuable insights \cite{li2013anomaly}.


\subsection{Training Strategy and Hyper-partameters}
\label{sec:train_param}
For all the training instances, we employed Adam Optimizer, and the training batch size was set to $256$ and $512$ for FTD and CTD branches respectively. As for all the training instances dropout rate and weight decay have been set to $0.1$ and $5e-5$ respectively. The training procedures were conducted on a workstation equipped with three NVIDIA RTX A6000 graphic cards and an AMD EPYC 7513 32-core processor. A conventional grid hyper-parameter search was systematically utilized to find the optimal set of hyper-parameters. 

\textbf{SHT} \cite{liu2018future} has been recorded at 24 FPS. Thus, we consider the input sequence length to be 24, equivalent to 1s. SPARTA-C underwent training for $20$ epochs with a learning rate of $1e-5$. In the next step, we freeze the parameters of the trained pose encoder and train the SPARTA-F decoder for $30$ epochs with a learning rate of $2e-3$. 

\textbf{HR-SHT} \cite{morais2019learning} contains the same videos in the training set as SHT. Thus, we do not have a separate training for it. We use the model trained on SHT and validate it on the HR-SHT subset as well.

\textbf{CHAD} \cite{danesh2023chad} is recorded at 30 FPS. Thus, we chose the input sequence length to be 1s or 30 frames. SPARTA-C is trained for $30$ epochs with a learning rate of $2e-3$. In the next step, We freeze the parameters of the pose encoder and train the SPARTA-F decoder for $30$ epochs with a learning rate of $5e-4$. 

\textbf{NWPUC} \cite{cao2023new} is recorded at 25 FPS. Thus, we chose the input sequence length to be 24 to be close to 1s as possible. SPARTA-C is trained for $30$ epochs with a learning rate of $5e-3$. In the next step, We freeze the parameters of the pose encoder and train the SPARTA-F decoder for $30$ epochs with a learning rate of $1e-3$. 

The training approach is entirely self-supervised, identifying the best model through the minimization of MSE loss on the training data. This model is then subjected to a single evaluation on the test set to assess its anomaly detection efficacy.


\section{Results}
\label{sec:results}

\subsection{Comparison With Pose-based Approaches}

\begin{figure*}[ht!]
    \centering
    \resizebox{0.75\linewidth}{!}{
    \includegraphics[clip,trim={5 5 0 5},width=1\textwidth]{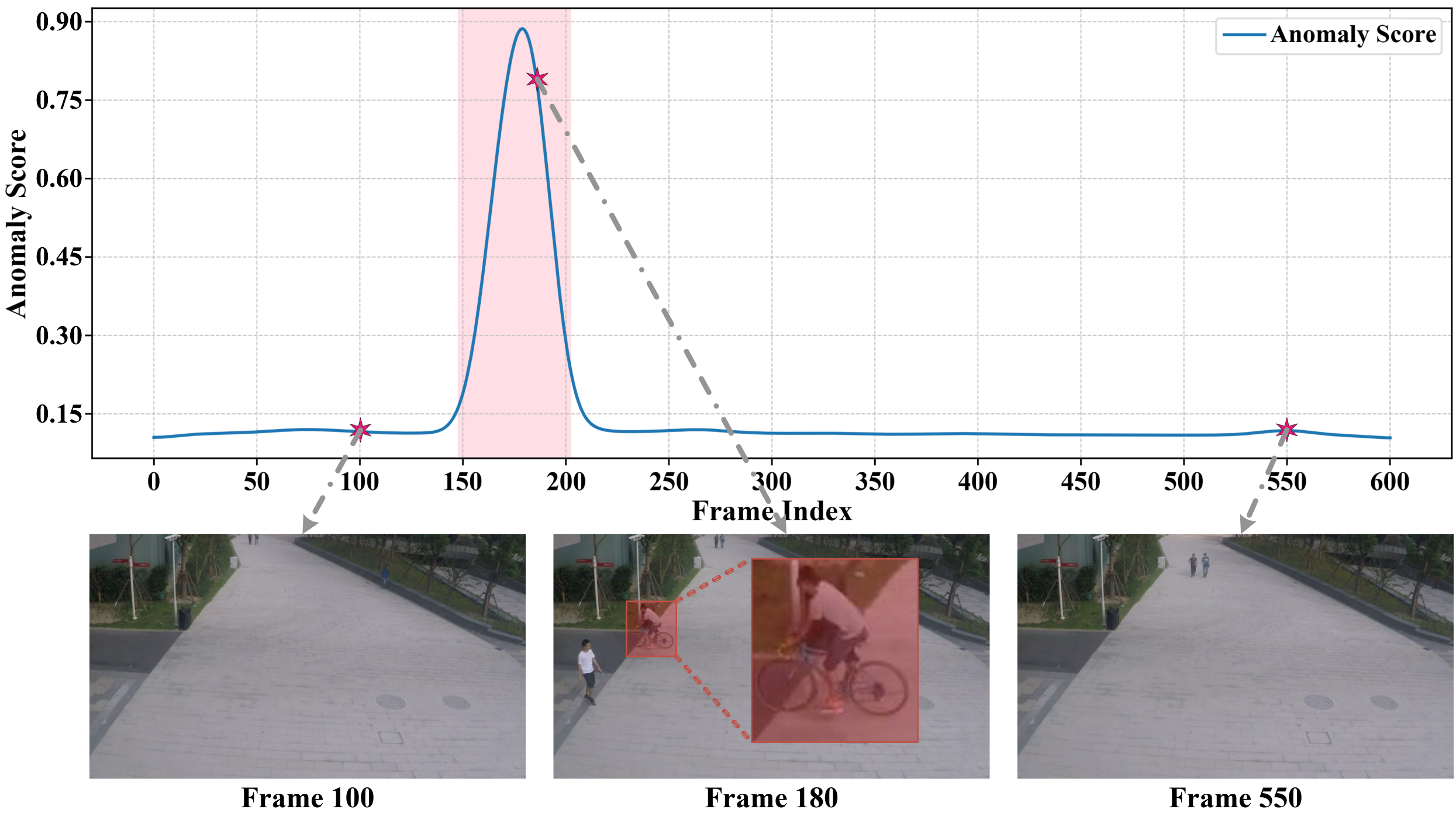}
    }
    \caption{Output anomaly scores of SPARTA-H for each frame of clip $01\_0025$ from the SHT dataset \cite{liu2018future}. The red area on the plot indicates the ground truth anomalous frames. In this clip, the anomalous behavior is a person riding a bike on the sidewalk, shown by the red rectangle.}
    \label{fig:shang_q}
\end{figure*}

\begin{table}[htbp]
\centering
\caption{AUC-ROC of SPARTA compared with pose-based approaches on SHT \cite{liu2018future}, HR-SHT \cite{morais2019learning}, CHAD \cite{danesh2023chad}, and NWPUC \cite{cao2023new} datasets. SPARTA is compared to SOTA methods such as MPED-RNN \cite{morais2019learning}, GEPC \cite{markovitz2020graph}, PoseCVAE \cite{jain2021posecvae}, MSTA-GCN \cite{chen2023multiscale}, MTP \cite{rodrigues2020multi}, HSTGCNN \cite{zeng2021hierarchical}, STGformer \cite{huang2022hierarchical}, MoPRL \cite{yu2023regularity} and STG-NF \cite{hirschorn2023normalizing}. The best is in bold and the second best is underlined.}
\label{tab:auc_roc}
\footnotesize
\begin{tabular}{l|ccccc}
\toprule[\heavyrulewidth] \midrule
\textbf{Methods}     & \textbf{SHT} & \textbf{HR-SHT} & \textbf{CHAD} & \textbf{NWPUC} & \textbf{Average} \\ \midrule
\textbf{MPED-RNN }    & 73.40        & 75.40           & -        & -          & - \\
\textbf{GEPC}        & 75.50        & -               & 64.90    & 62.04          & - \\
\textbf{PoseCVAE}    & 74.90        & 75.70           & -        & -          & - \\
\textbf{MSTA-GCN }    & 75.90        & -               & -        & -          & - \\
\textbf{MTP}         & 76.03        & 77.04           & -        & -          & - \\
\textbf{HSTGCNN}     & 81.80        & 83.40           & -        & -          & - \\
\textbf{STGformer}   & 82.90        & 86.97           & -        & -          & - \\
\textbf{MoPRL}       & 83.35        & 84.40           & 
\underline{66.81}    & 61.92          &  74.12 \\ 
\textbf{STG-NF}      & \textbf{85.90} & \textbf{87.40} & 60.60    & 62.56          &  74.11 \\ \midrule
\textbf{SPARTA-C} & 85.10        & 86.70           & 66.12 & \underline{62.69}          & \underline{75.15} \\
\textbf{SPARTA-F} & 83.19        & 83.70           & 66.61    & 62.29          &  73.94\\
\textbf{SPARTA-H} & \underline{85.75} & \underline{87.23} & \textbf{67.04} & \textbf{63.48}          & \textbf{75.87} \\ \midrule \bottomrule[\heavyrulewidth]
\end{tabular}
\end{table}

\begin{table}[htbp]
\centering
\caption{EER of SPARTA compared with pose-based approaches on SHT \cite{liu2018future}, HR-SHT \cite{morais2019learning}, CHAD \cite{danesh2023chad}, and NWPUC \cite{cao2023new} datasets. SPARTA is compared to SOTA methods such as GEPC \cite{markovitz2020graph}, STG-NF \cite{hirschorn2023normalizing}, and MoPRL \cite{yu2023regularity}. The best is in bold and the second best is underlined.}
\label{tab:eer}
\footnotesize
\begin{tabular}{l|ccccc}
\toprule[\heavyrulewidth] \midrule
\textbf{Methods}     & \textbf{SHT} & \textbf{HR-SHT} & \textbf{CHAD} & \textbf{NWPUC} & \textbf{Average} \\ \midrule
\textbf{GEPC}        & 0.31        & -               & \underline{0.38}       & 0.41               & - \\
\textbf{STG-NF}      & \textbf{0.22} & \textbf{0.21} & 0.43            &\underline{0.40}              & \underline{0.31} \\ 
\textbf{MoPRL}      & 0.24 & 0.23 & \underline{0.38} & \underline{0.40} &  \underline{0.31}\\ \midrule
\textbf{SPARTA-C} & \underline{0.23} & \underline{0.22} & \underline{0.38} & 0.41 & \underline{0.31} \\
\textbf{SPARTA-F} & 0.25        & 0.25           & \underline{0.38}  & \underline{0.40}              & 0.32 \\
\textbf{SPARTA-H} & \textbf{0.22} & \textbf{0.21} & \textbf{0.37} & \textbf{0.39}             &  \textbf{0.29}\\ \midrule \bottomrule[\heavyrulewidth]
\end{tabular}
\end{table}

SPARTA-H achieves the highest average AUC-ROC, surpassing the previous SOTA by $1.75\%$ across four benchmark datasets (\Cref{tab:auc_roc}). It outperforms the prior best model by $0.23\%$ and $0.92\%$ on CHAD and NWPUC, respectively, and ranks second on SHT and HR-SHT, trailing STG-NF by only $0.15\%$ and $0.17\%$, whose reliance on a fixed normal distribution limits its generalizability, particularly on other diverse datasets. These results highlight SPARTA-H's robustness and versatility, making it the most effective model overall.

Examining the EER across all datasets (\Cref{tab:eer}), SPARTA-H achieves the lowest average EER of $0.29$, outperforming previous SOTA models. This aligns with the AUC-ROC results, reaffirming SPARTA-H's superior performance and generalizability. SPARTA-H matches STG-NF on SHT and HR-SHT but demonstrates better robustness on CHAD and NWPUC with lower EER values, highlighting its versatility and effectiveness for real-world applications that require a balance between usability and security.


As shown in \Cref{tab:auc_roc}, SPARTA-H outperforms its variants, SPARTA-C and SPARTA-F, consistently across datasets. For instance, on the SHT dataset, SPARTA-H achieves an AUC-ROC of $85.75\%$, compared to SPARTA-C's $85.10\%$ and SPARTA-F's $83.19\%$, demonstrating the effectiveness of its dual-branch design. The complementary synergy of CTD and FTD branches allows SPARTA-H to detect anomalies that might be missed by either branch alone. This advantage extends to EER (\Cref{tab:eer}), further underscoring the robustness and importance of the dual-branch architecture, which enhances performance and generalizability.

\begin{figure*}[ht!]
    \centering
    \resizebox{0.75\linewidth}{!}{
    \includegraphics[clip,trim={5 5 5 5},width=1\textwidth]{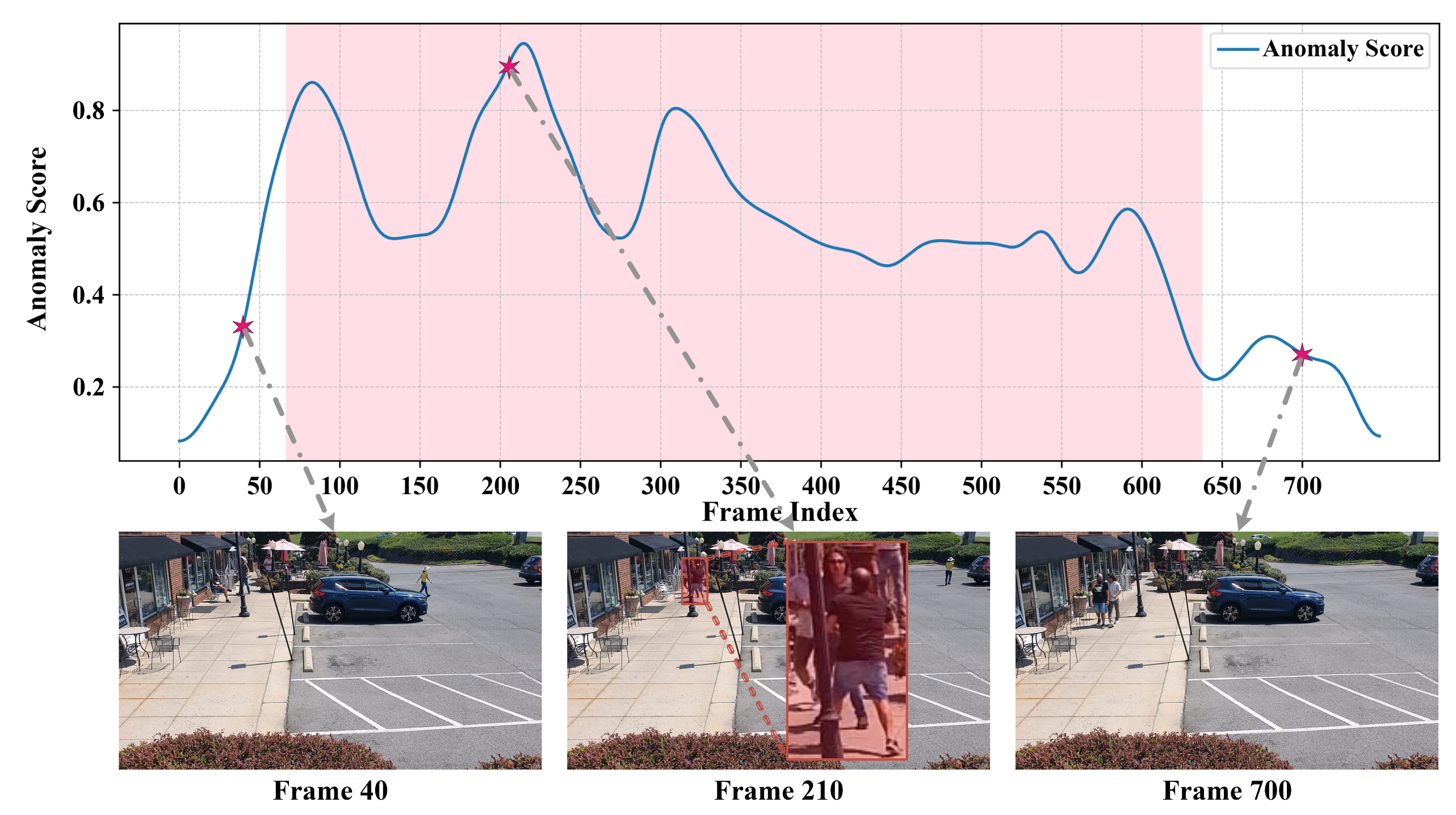}
    }
    \caption{Output anomaly scores of SPARTA-H for each frame of clip $04\_093\_1$ from the CHAD dataset \cite{danesh2023chad}. The red area on the plot indicates the ground truth anomalous frames. In this clip, the anomalous behavior is two people fighting shown by the red rectangle.}
    \label{fig:chad_q}
\end{figure*}


\begin{table}[htbp]
\centering
\caption{AUC-ROC of SPARTA compared with pixel-based approaches on SHT \cite{liu2018future} and NWPUC \cite{cao2023new} datasets. SPARTA is compared with SOTA methods such as MNAD \cite{park2020learning}, OG-Net \cite{zaheer2020old}, MemAE \cite{gong2019memorizing}, MAAM-Net \cite{wang2023memory}, MPN \cite{lv2021learning}, LLSH \cite{lu2022learnable}, GCL$_{PT}$ \cite{zaheer2022generative}, BAF \cite{georgescu2021background}, BAF \cite{georgescu2021background} + SSPCAB \cite{ristea2022self}, SSMTL \cite{georgescu2021anomaly}, SSMTL++v2 \cite{barbalau2023ssmtl++}, Jigsaw-VAD \cite{wang2022video} and NM-GAN \cite{chen2021nm}. The best is in bold and the second best is underlined.}
\label{tab:pixel}
\begin{tabular}{l|ccc}
\toprule[\heavyrulewidth] \midrule
\textbf{Methods}     & \textbf{SHT} & \textbf{NWPUC} & \textbf{Average} \\ \midrule
\textbf{MNAD}   & 70.50       & 62.50      & 66.50 \\
\textbf{OG-Net}   & -           & 62.50      & -      \\
\textbf{MemAE}   & 71.20        & 61.90      & 66.55 \\
\textbf{MAAM-Net}   & 71.30        & -      & -      \\
\textbf{MPN}   & 73.80       & \textbf{64.40}     & 69.10 \\
\textbf{LLSH}   & 77.60       & 62.20      & 69.90 \\
\textbf{GCL$_{\textbf{PT}}$ }   & 78.93   & -      & -      \\
\textbf{BAF}   & 82.70        & -      & -      \\
\textbf{BAF + SSPCAB }   & 83.60   & -      & -      \\
\textbf{SSMTL}       & 83.50        & -      & -      \\ 
\textbf{SSMTL++v2}       & 83.80        & -      & -      \\ 
\textbf{Jigsaw-VAD}       & 84.30       & -      & -      \\ 
\textbf{NM-GAN}   & \underline{85.30} & -      & -      \\
\midrule
\textbf{SPARTA-C} & 85.10        & 62.69  & \underline{73.90} \\
\textbf{SPARTA-F} & 83.19        & 62.29  & 72.74 \\
\textbf{SPARTA-H} & \textbf{85.75}  & \underline{63.48} & \textbf{74.62} \\ \midrule \bottomrule[\heavyrulewidth]
\end{tabular}%
\end{table}

To further illustrate the effectiveness of the proposed model, two examples from SHT \cite{liu2018future} and CHAD \cite{danesh2023chad} are shown in \Cref{fig:shang_q} and \Cref{fig:chad_q}. The y-axis represents the scaled output anomaly score of SPARTA-H for all frames of the testing clip. As depicted in \Cref{fig:shang_q}, SPARTA-H accurately detects anomalous behavior, maintaining a steady score on normal frames and showing an increased score on anomalous ones. \Cref{fig:chad_q} similarly demonstrates that SPARTA-H produces higher scores for anomalous frames, enabling anomaly detection. However, more noise is evident in \Cref{fig:chad_q}, and it is not as accurate as in \Cref{fig:shang_q}, which is also reflected by the lower AUC-ROC and higher EER observed on the CHAD dataset compared to the SHT dataset showcased in \Cref{tab:auc_roc} and \Cref{tab:eer}. 

\subsection{Comparison With Pixel-based Approaches}

This manuscript mainly focuses on pose-based approaches. However, we will further explore an additional comparative analysis between SPARTA and pixel-based methodologies to have a better understanding of SPARTA's capabilities. The datasets common between pose-based and pixel-based approaches are SHT \cite{liu2018future} and NWPUC \cite{cao2023new}. Notably, CHAD \cite{danesh2023chad} has not yet been employed in pixel-based studies. Therefore, we further compare SPARTA with SOTA pixel-based algorithms on SHT and NWPUC.





\begin{table}[]
\centering
\caption{Evaluating AUC-ROC on SHT \cite{liu2018future}, HR-SHT \cite{morais2019learning}, CHAD \cite{danesh2023chad}, and NWPUC \cite{cao2023new} datasets: A comparative analysis of our design variants with and without incorporating relative motion. The best result of each branch is highlighted in gray.}
\footnotesize
\label{tab:relative}
\begin{tabular}{l|c|cccc}
\toprule[\heavyrulewidth] \midrule
& \makecell{\textbf{Relative} \\ \textbf{Movement}}   & \textbf{SHT} & \textbf{HR-SHT} & \textbf{CHAD} & \textbf{NWPUC} \\ \midrule

\multirow{2}{*}{\textbf{SPARTA-C}} 
 & \xmark & 82.97 & 84.80 & 57.56 & 62.52 \\
 & \cmark & \cellcolor[HTML]{EFEFEF}85.10 & \cellcolor[HTML]{EFEFEF}86.70 & \cellcolor[HTML]{EFEFEF}66.12 &  \cellcolor[HTML]{EFEFEF}62.69 \\ \hline

\multirow{2}{*}{\textbf{SPARTA-F}} 
 & \xmark & 81.80 & 82.80 & 58.27 &  \cellcolor[HTML]{EFEFEF}62.72 \\
 & \cmark & \cellcolor[HTML]{EFEFEF}83.19 & \cellcolor[HTML]{EFEFEF}83.70 & \cellcolor[HTML]{EFEFEF}66.61 & 62.29 \\ \hline

\multirow{2}{*}{\textbf{SPARTA-H}} 
 & \xmark & 84.20 & 85.47 & 57.95 & 63.41 \\
 & \cmark & \cellcolor[HTML]{EFEFEF}85.75 & \cellcolor[HTML]{EFEFEF}87.23 & \cellcolor[HTML]{EFEFEF}67.04 & \cellcolor[HTML]{EFEFEF} 63.48 \\

\midrule \bottomrule[\heavyrulewidth]
\end{tabular}
\end{table}

\begin{figure*}[ht!]
    \centering
    \includegraphics[clip,trim={18 18 18 18},width=0.85\textwidth]{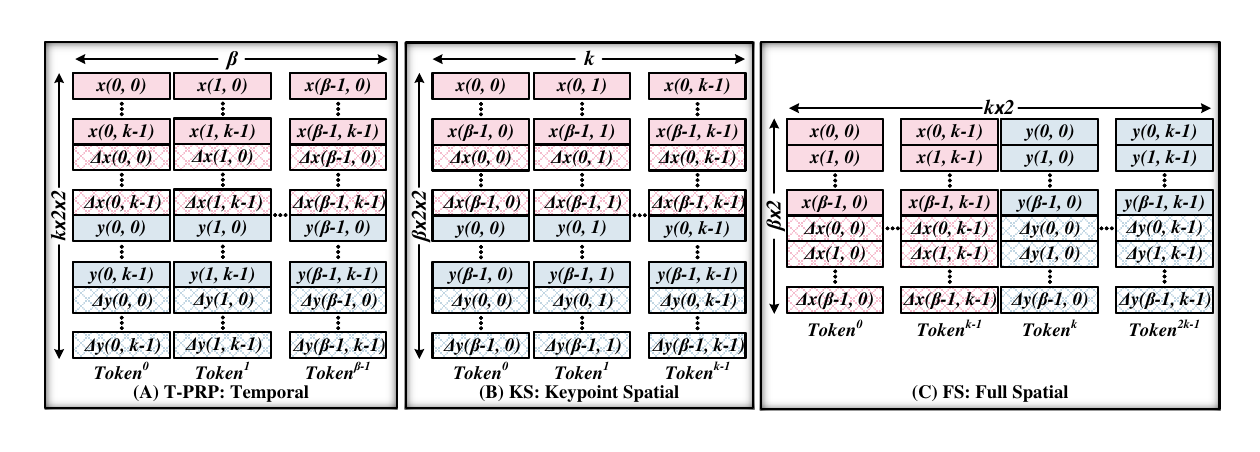}
    \caption{Proposed tokenization methods. $k$ is the number of keypoints, $\beta$ is the input window size, $\Delta$ shows relative coordinates and $x(t, k)$ and $y(t, k)$ are the coordinates of $k^{th}$ keypoint in time step $t$.}
    \label{fig:tokens}
\end{figure*}

As outlined in \Cref{sec:intro} and \Cref{sec:related}, pixel-based approaches have historically been regarded as more precise than pose-based approaches. Nonetheless, recent studies indicate a shift in this trend. In \Cref{tab:pixel}, we present a comparative analysis of SPARTA against current SOTA pixel-based algorithms. SPARTA-H achieves an average AUC-ROC of $74.62\%$ across the SHT and NWPUC datasets, underscoring its overall superiority. Pose-based approaches inherently exhibits lower bias and reduced susceptibility to background noise, while simultaneously promoting greater privacy and adhering to ethical standards \cite{alinezhad2023understanding, ardabili2023understanding, noghre2024exploratory}.

\section{Ablation Study}
\label{sec:ablation}
\subsection{Impact of Relative Pose}
\label{sec:rel}

The results detailed in \Cref{tab:relative} empirically validate the effectiveness of using relative pose, as theoretically outlined in \Cref{sec:st-prp}. This empirical evidence complements and reinforces the theory that incorporating relative movement benefits anomaly detection. The results indicate a consistent improvement across all variations of the SPARTA when the relative pose is integrated. Specifically, regarding AUC-ROC, the SPARTA-H variant incorporating relative movement demonstrates a notable performance enhancement. On the SHT dataset, incorporating relative movement in SPARTA-H improves AUC-ROC by $1.55\%$, demonstrating its effectiveness in capturing motion dynamics. This effect is even more pronounced on CHAD, with a $9.09\%$ improvement, highlighting the significance of relative pose in complex scenarios. On NWPUC \cite{cao2023new}, SPARTA-C and SPARTA-H exhibit similar gains, while SPARTA-F shows a slight decrease, suggesting dataset-specific variations. Overall, these results affirm that relative movement generally provides complementary information and enhances anomaly detection performance.

\begin{table}[]
\centering
\caption{Evaluating AUC-ROC on SHT \cite{liu2018future}, HR-SHT \cite{morais2019learning}, CHAD \cite{danesh2023chad}, and NWPUC \cite{cao2023new} datasets: A comparative analysis of our design with different methods of tokenization. The best result of each branch is highlighted in gray.}
\label{tab:tokenization}
\resizebox{1\columnwidth}{!}{%
\begin{tabular}{l|c|cccc}
\toprule[\heavyrulewidth] \midrule
&  \textbf{Tokenization}   & \begin{tabular}[c]{@{}c@{}}\textbf{SHT}\end{tabular} & \begin{tabular}[c]{@{}c@{}}\textbf{HR-SHT}\end{tabular} & \begin{tabular}[c]{@{}c@{}}\textbf{CHAD}\end{tabular} & \begin{tabular}[c]{@{}c@{}}\textbf{NWPUC}\end{tabular} \\ \midrule

\multirow{4}{*}{\textbf{SPARTA-C }} & T-PRP &   83.85     &      85.46     &     65.59   & 59.93 \\
 & KS-PRP &   83.67     &      85.15     &     64.69     & 62.67 \\ 
 & FS-PRP &    83.45    &     85.13      &     62.84     & 61.96 \\ 
 & ST-PRP & \cellcolor[HTML]{EFEFEF}85.10        & \cellcolor[HTML]{EFEFEF}86.70           & \cellcolor[HTML]{EFEFEF}66.12         & \cellcolor[HTML]{EFEFEF}62.69 \\ \hline
\multirow{4}{*}{\textbf{SPARTA-F }} & T-PRP &    82.08   &    \cellcolor[HTML]{EFEFEF}84.47       &     66.18   & 61.85 \\
 & KS-PRP &    79.53    &     80.40      &       64.95   &  61.91\\ 
 & FS-PRP &    82.08    &      83.21     &      64.29    & 62.00 \\ 
 & ST-PRP &    \cellcolor[HTML]{EFEFEF}83.19   &       83.70   &    \cellcolor[HTML]{EFEFEF}66.61 &\cellcolor[HTML]{EFEFEF}62.29 \\ \hline
\multirow{4}{*}{\textbf{SPARTA-H }} & T-PRP &   84.40     &    86.06     & \cellcolor[HTML]{EFEFEF}67.06    & 61.95  \\
 & KS-PRP &    83.63    &     84.94      &     65.37     & 62.67 \\ 
 & FS-PRP &    84.68    &     86.37      &     64.28     &  62.45\\ 
 & ST-PRP &    \cellcolor[HTML]{EFEFEF}85.75   &    \cellcolor[HTML]{EFEFEF}87.23       &    67.04   & \cellcolor[HTML]{EFEFEF}63.48 \\
\midrule \bottomrule[\heavyrulewidth]
\end{tabular}%
}
\end{table}

\begin{figure*}[ht!]
    \centering
    \includegraphics[clip,trim={5 5 5 5},width=1\textwidth]{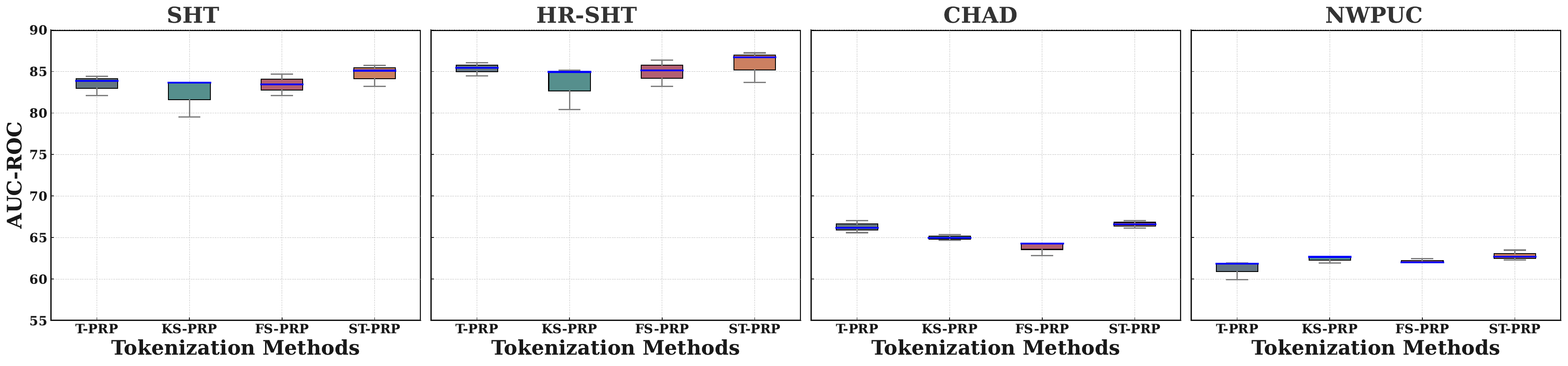}
    \caption{Box-and-whisker plots of tokenization methods' AUC-ROC performance on SHT \cite{liu2018future}, HR-SHT \cite{morais2019learning}, CHAD \cite{danesh2023chad}, and NWPUC \cite{cao2023new} datasets.}
    \label{fig:tokens_roc}
\end{figure*}
\subsection{Exploring Diverse Tokenization Strategies}
\label{sec:tokenization}

We employed diverse tokenization strategies to optimize the synergy between temporal and spatial attention. This experiment aims to identify the most effective method for the transformer core to interpret and analyze pose behavior more accurately, optimizing its overall anomaly detection capabilities. The input window of size $\beta$ and the number of keypoints $k$ remain consistent across all implemented strategies, ensuring a uniform basis for comparability. For each tokenization method, the best model was selected after a grid hyperparameter search; details can be found in the supplementary materials.

\textbf{Temporal Pose and Relative Pose (T-PRP)} tokenization prioritizes the temporal motion between video frames by encapsulating the information of an individual frame, encompassing its pose and relative pose represented as $(x, y)$ and $(\Delta x, \Delta y)$ coordinates, within a single token - underscoring the sequential nature of the frames. As depicted in \Cref{fig:tokens} part (A) each token has dimensions of $k \times 2 \times 2$ and the number of tokens matches the window size $\beta$.

\textbf{Keypoint Spatial Pose and Relative Pose (KS-PRP)} tokenization focuses on the interrelation among keypoints within a sequence of poses. As illustrated in \Cref{fig:tokens} (B), each token encapsulates the positional information of a specific keypoint (e.g., elbow) in terms of $x$ and $y$ coordinates across all frames within the input window. Consequently, the size of each token is $\beta \times 2 \times 2$. This creates $k$ input tokens, each representing one of the $k^{th}$ keypoints.

\textbf{Full Spatial Pose and Relative Pose (FS-PRP)} tokenization, similar to KP tokenization, also focuses on the relationship between the keypoints of a pose, it takes into account that there is a relationship between the $x$ and $y$ of a certain keypoint (e.g. elbow) too. Consequently, FS tokenization refines the tokenization scheme of KS tokenization further by partitioning $x$ and $y$ coordinates, thereby creating $k \times 2$ tokens. The initial set of $k$ tokens pertains to $x$ coordinates, while the subsequent set of $k$ tokens pertains to $y$ coordinates.

On top of all these tokenization strategies, we also add positional encoding to embed order into input sequences. In \Cref{tab:tokenization}, the performance of various tokenization methods is compared. The ST-PRP tokenization method, detailed in \Cref{sec:st-prp}, demonstrates superior performance in most cases compared to other approaches. While the T-PRP tokenization outperforms others in two specific instances, purely spatial tokenizations (KS-PRP and FS-PRP) consistently yield suboptimal results. \Cref{fig:tokens_roc} further supports this observation, clearly demonstrating that across all four benchmark datasets, the SPARTA variants achieve higher overall AUC-ROC when using ST-PRP tokenization. This suggests that the combination of temporal and spatial attention between tokens uncovers crucial information for analyzing human behavior patterns that neither can detect independently. Consequently, the ST-PRP tokenization method, which integrates spatial and temporal information, emerges as the most effective approach.



\section{Conclusion}
\label{sec:conclusion}

In this paper, we introduced methodologies that pave the way for advanced human-centric VAD. The novel proposed Spatio-Temporal Pose and Relative Pose (ST-PRP) tokenization method, which serves as a key component for high-level human behavior analysis. Combined with our new Unified Encoder Twin Decoders (UETD) transformer core, the proposed SPARTA architecture demonstrates superior performance in self-supervised human-centric VAD. Extensive benchmarking against SOTA methods confirms SPARTA's accuracy and robustness. We hope that our contributions will serve as a foundation for future advancements in the field.

\section*{Acknowledgments}
This research is supported by the National Science Foundation (NSF) under Award Numbers 1831795 and 2329816.


{\appendices
\section*{Appendix: Ablation Studies Hyperparameters}\label{appendix}
This section provides an in-depth exposition of the architectural and training hyperparameters of the ablation studies (\Cref{sec:ablation}) to ensure the reproducibility of the results. 

\subsection{Relative Pose Ablation Setup and Hyperparameters}
To reveal the benefit of incorporating relative movement, we use the best SPARTA model which includes 12 heads and 4 layers with the feed-forward layer size set to 64. This variant is trained with and without relative movement data using the hyperparameters shown in \Cref{tab:train_rel}. The optimal hyperparameters are chosen using a systematic grid search. In all training instances, we have used the Adam optimizer with a weight decay of $5.0e-5$ and trained branches for 30 epochs. For the training, the same strategy is used as in \Cref{sec:SPARTA}; first, the SPARTA-C (the unified encoder and CTD decoder) is trained. In the next step, the unified encoder is frozen and the FTD decoder is trained. Since both the SHT and HR-SHT utilize identical videos in their training sets, the hyperparameters for both models remain consistent.

\begin{table}[]
\centering

\caption{The training hyperparameters used for the Relative Pose Ablation Study (\Cref{sec:rel}) on SHT \cite{liu2018future}, HR-SHT \cite{morais2019learning}, CHAD \cite{danesh2023chad}, and NWPUC \cite{cao2023new} datasets. LR and DR refer to the Learning Rate and Dropout Rate.}
\label{tab:train_rel}
\resizebox{\columnwidth}{!}{%
\begin{tabular}{@{}c|c|cc|cc|cc@{}}
\toprule
\toprule
                             &                                                             & \multicolumn{2}{c|}{\begin{tabular}[c]{@{}c@{}}\textbf{SHT,}\\ \textbf{HR-SHT}\end{tabular}} & \multicolumn{2}{c|}{\textbf{CHAD}} & \multicolumn{2}{c}{\textbf{NWPUC}} \\ \midrule
                             & \begin{tabular}[c]{@{}c@{}}\textbf{Relative}\\ \textbf{Movement}\end{tabular} & \textbf{LR}                                    & \textbf{DR}                              & \textbf{LR}          & \textbf{DR}    & \textbf{LR} & \textbf{DR} \\ \midrule
\multirow{2}{*}{\textbf{SPARTA-C}} & \xmark                                                          & $2.0e-3$                               & $0.1$                                  & $5.0e-6$     & $0.1$        & $1.0e-2$ & $0.2$ \\
                             & \cmark                                                         & $1.0e-5$                               & $0.1$                                  & $2.0e-3$     & $0.1$        & $5.0e-3$  & $0.3$ \\
                             \midrule
\multirow{2}{*}{\textbf{SPARTA-F}} & \xmark                                                          & $5.0e-4$                               & $0.1$                                  & $3.0e-3$     & $0.1$        & $1.0e-3$ & $0.1$ \\
                             & \cmark                                                         & $2.0e-3$                               & $0.1$                                  & $5.0e-4$     & $0.1$        & $1.0e-3$ & $0.1$ \\ \bottomrule \bottomrule
\end{tabular}%
}
\end{table}

\section*{Appendix: Tokeniation Ablation Setup and Hyperparamters}
To effectively compare various tokenization setups at their full potential, we carried out a systematic grid search. This approach was not just to identify the optimal training hyperparameters, but also to determine the best architectural choices for each setup. 

\begin{table}[]
\centering
\caption{The design choices used for the Tokenization Ablation 
Study (\Cref{sec:tokenization})}
\label{tab:arch_param}
\begin{tabular}{@{}c|ccc@{}}
\toprule
\toprule
                & \textbf{\# Heads} & \textbf{\# Layers} & \textbf{\begin{tabular}[c]{@{}c@{}}Feed Forward\\ Dimension\end{tabular}} \\ \midrule
\textbf{T-PRP}  & 8                & 8                 & 128                                                                       \\
\textbf{KS-PRP} & 8                & 4                 & 128                                                                       \\
\textbf{FS-PRP} & 12               & 6                 & 64                                                                        \\
\textbf{ST-PRP} & 12               & 4                 & 64                                                                        \\ \bottomrule \bottomrule
\end{tabular}%
\end{table}

\begin{table}[]
\centering
\caption{The training hyperparameters used for the Tokenization Ablation Study (\Cref{sec:tokenization}) on SHT \cite{liu2018future}, HR-SHT \cite{morais2019learning}, CHAD \cite{danesh2023chad}, and NWPUC \cite{cao2023new} datasets. LR and DR refer to the Learning Rate and Dropout Rate.}
\label{tab:table_param}
\resizebox{\columnwidth}{!}{%
\begin{tabular}{@{}c|c|cc|cc|cc@{}}
\toprule
\toprule
\multicolumn{1}{l}{}                  & \multicolumn{1}{l|}{} & \multicolumn{2}{c|}{\textbf{\begin{tabular}[c]{@{}c@{}}SHT, \\ HR-SHT \end{tabular}}} & \multicolumn{2}{c|}{\textbf{CHAD }} & \multicolumn{2}{c}{\textbf{NWPUC}} \\ \midrule
\multicolumn{1}{l}{}                  & \multicolumn{1}{l|}{} & \textbf{LR}                             & \textbf{DR}                            & \textbf{LR}   & \textbf{DR}  & \textbf{LR}   & \textbf{DR}  \\ \midrule
\multirow{4}{*}{\textbf{SPARTA-C}} & \textbf{T-PRP}       & $5.0e-6$                                & $0.1$                                         & $3.0e-3$      & $0.1$        &         $1.0e-3$     & $0.1$            \\
                                      & \textbf{KS-PRP}      & $5.0e-6$                                & $0.1$                                         & $3.0e-3$      & $0.2$        & $1.0e-3$             & $0.1$            \\
                                      & \textbf{FS-PRP}      & $1.0e-5$                                & $0.1$                                         & $1.0e-3$      & $0.1$        & $1.0e-3$             & $0.1$            \\
                                      & \textbf{ST-PRP}      & $1.0e-5$                                & $0.1$                                         & $2.0e-3$      & $0.1$        & $5.0e-3$             & $0.1$            \\ \midrule
\multirow{4}{*}{\textbf{SPARTA-F}} & \textbf{T-PRP}       & $3.0e-3$                                & $0.1$                                         & $1.0e-4$      & $0.1$        & $2.0e-3$            & $0.1$            \\
                                      & \textbf{KS-PRP}      & $1.0e-4$                                & $0.2$                                         & $1.0e-4$      & $0.1$        & $3.0e-3$              & $0.1$            \\
                                      & \textbf{FS-PRP}      & $5.0e-4$                                & $0.1$                                         & $1.0e-5$      & $0.1$        & $1.0e-3$           & $0.1$           \\
                                      & \textbf{ST-PRP}      & $2.0e-3$                                & $0.1$                                         & $5.0e-4$      & $0.1$        & $1.0e-3$             & $0.1$            \\ \bottomrule \bottomrule
\end{tabular}%
}
\end{table}

\Cref{tab:arch_param} presents the architectural design parameter choices obtained from the grid search on SHT dataset \cite{liu2018future}. These parameters are kept the same for other datasets to ensure a fair comparison. Across all experiments, both the unified encoder and the twin decoders consistently adhere to the parameters detailed in \Cref{tab:arch_param}. This consistency ensures a standardized approach in our experimental setup. On the other hand, \Cref{tab:table_param} shows the best learning rate and dropout for training per branch. The number of epochs, optimizer, and weight decay are 30, Adam, and $5.0e-5$ respectively.

Regarding the training process, same as other tests, SPARTA-C is initially trained following the strategy outlined in \cref{sec:SPARTA}, where both the unified encoder and the CTD decoder are trained together. Subsequently, for SPARTA-F, the unified encoder is frozen, and then only the FTD decoder undergoes further training. As the training sets for both the SHT and HR-SHT consist of the same videos, the hyperparameters are kept uniform for them.}

%

\bibliographystyle{IEEEtran}
\bibliography{main.bib}

\newpage

 

\begin{IEEEbiography}[{\includegraphics[width=1in,height=1.25in,keepaspectratio]{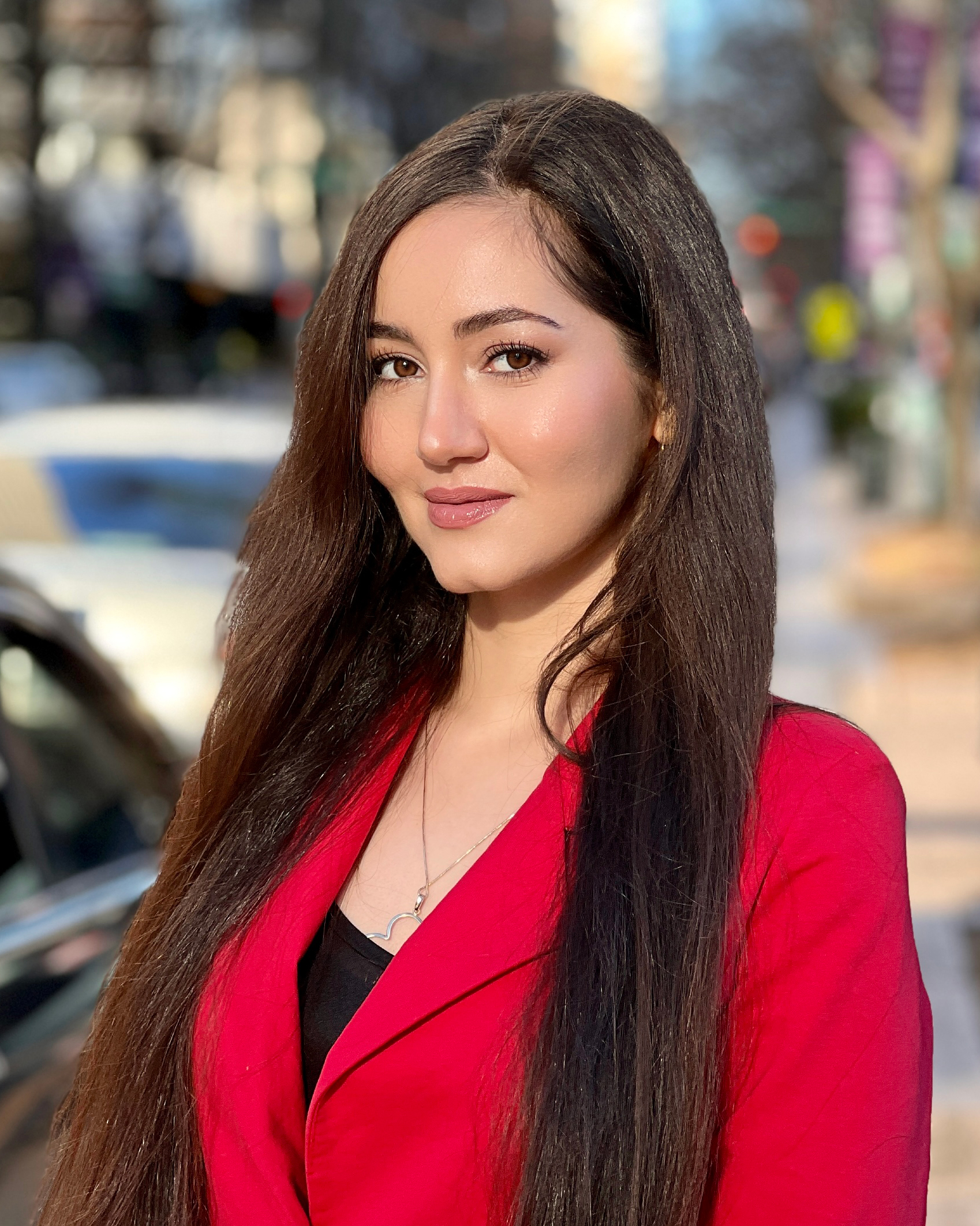}}]{Ghazal Alinezhad Noghre} (S’22) is currently a Ph.D. candidate in Electrical and Computer Engineering at the University of North Carolina at Charlotte, NC, United States. Her research concentrates on Artificial Intelligence, Machine Learning, and Computer Vision. She is particularly interested in the applications of anomaly detection, action recognition, and path prediction in real-world environments, and the challenges associated with these fields.
\end{IEEEbiography}

\begin{IEEEbiography}[{\includegraphics[width=1in,height=1.25in,keepaspectratio]{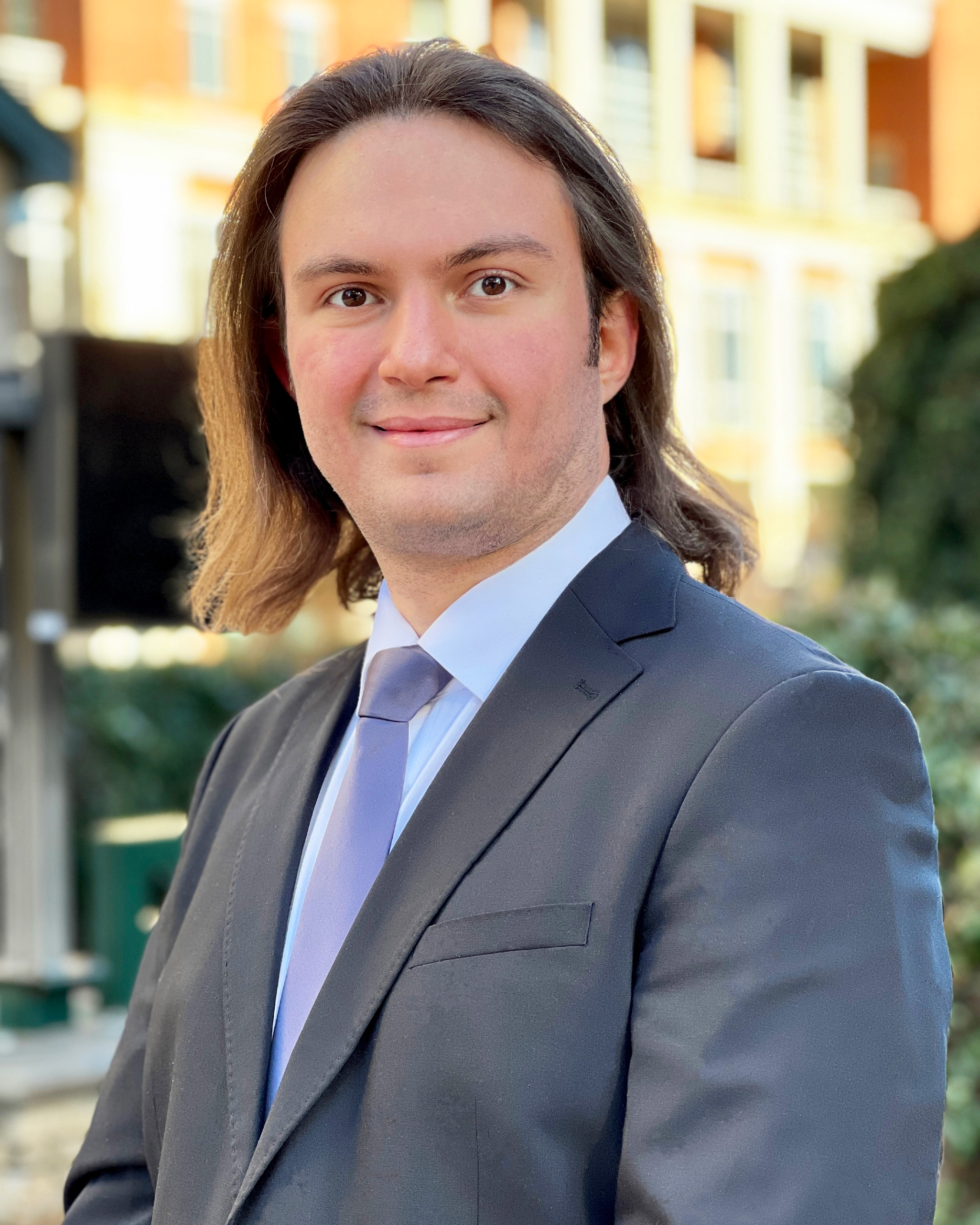}}]{Armin Danesh Pazho} (S’22) is currently a Ph.D. candidate at the University of North Carolina at Charlotte, NC, United States. With a focus on Artificial Intelligence, Computer Vision, and Deep Learning, his research delves into the realm of developing AI for practical, real-world applications and addressing the challenges and requirements inherent in these fields. Specifically, his research covers action recognition, anomaly detection, person re-identification, human pose estimation, and path prediction.
\end{IEEEbiography}

\begin{IEEEbiography}[{\includegraphics[width=1in,height=1.25in,keepaspectratio]{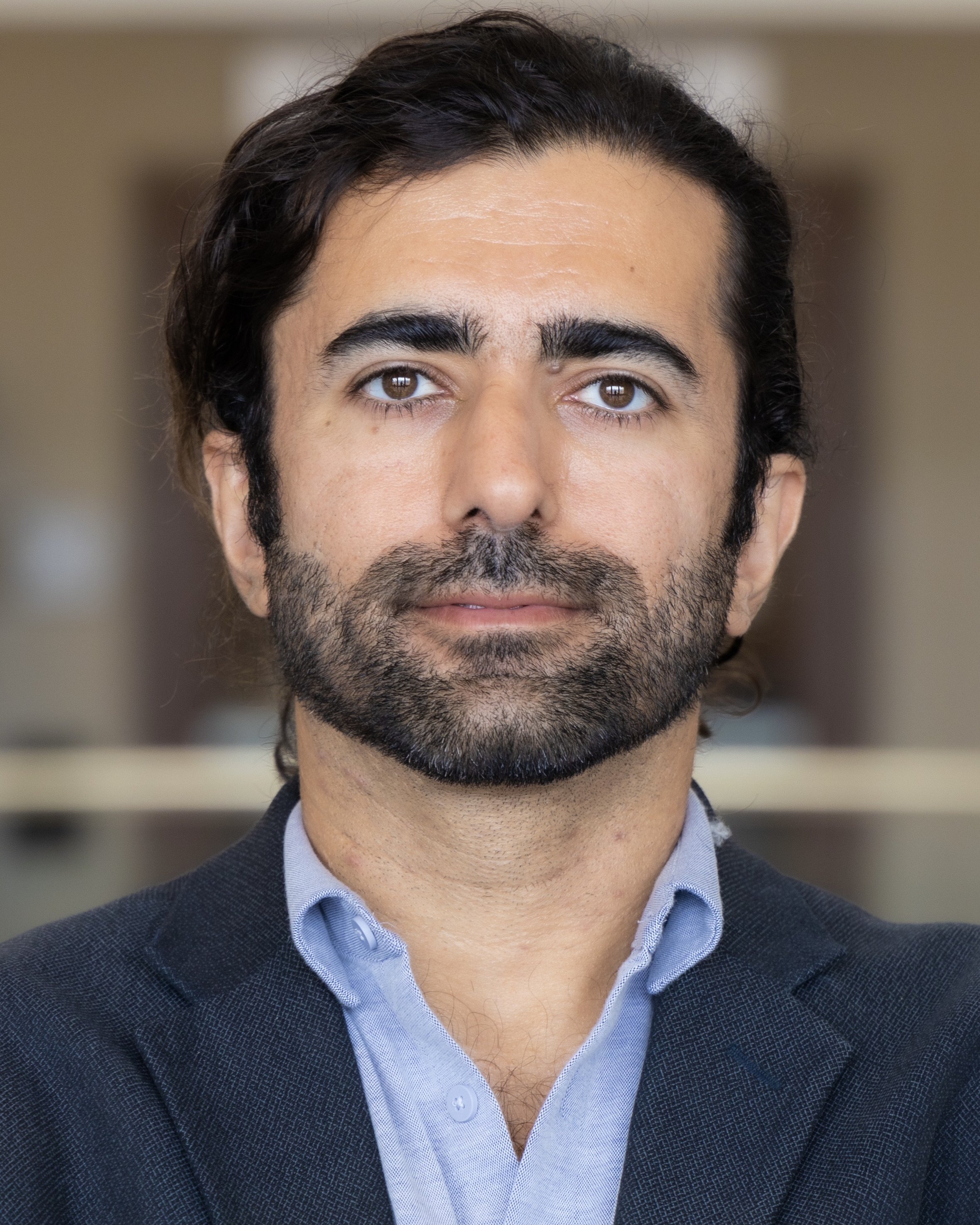}}]{Hamed Tabkhi} (S’07–M’14) is the associate professor of Computer Engineering at the University of North Carolina Charlotte (UNC Charlotte). He received his PhD in Computer Engineering from Northeastern University in 2014. His research and scholarship activities focus on transformative computer system solutions to bring recent advances in Artificial Intelligence (AI) to address real-world problems. In particular, he focuses on AI-based solutions to enhance our communities' safety, health, and overall well-being.
\end{IEEEbiography}

\vspace{11pt}


\vfill

\end{document}